\pdfoutput=1 %

\documentclass[11pt,a4paper]{article}
\usepackage[hyperref]{acl}
\usepackage[times]{stroop}

\title{$k$NN For Whisper And Its Effect On Bias And Speaker Adaptation}

\author{%
Maya K. Nachesa \qquad
Vlad Niculae \\ %
Language Technology Lab, University of Amsterdam\\
\{
\emldisplay{m.k.nachesa@uva.nl}{m.k.nachesa},~
\emldisplay{v.niculae@uva.nl}{v.niculae} %
\}\texttt{@uva.nl}
}
\date{}

\begin{document}
\maketitle
\begin{abstract}
Speech recognition performance varies by language, domain, and speaker characteristics such as accent, 
but
fine-tuning a model on any of these categories may lead to catastrophic forgetting. 
Token-level
$k$ nearest neighbor search ($k$NN), first proposed for neural sequence decoders for natural language generation (NLG) and machine translation (MT), is a non-parametric method that 
instead adapts using inference-time search in an external datastore,
without training the underlying model. We show that Whisper, a transformer end-to-end speech 
recognition
model, benefits from $k$NN. We investigate the differences between the speech and text setups. We discuss implications for speaker adaptation, and analyze improvements by gender, accent, and age.
\end{abstract}

\section{Introduction}
Automatic speech recognition (ASR) has improved significantly over the years. A recent success has been the end-to-end transformer encoder-decoder Whisper model \citep{radford2023whisper}. Besides its architecture, in the landscape of data scarcity that plagues speech recognition, Whisper stands out in having been trained on over 680,000 hours of transcribed audio on a wide variety of languages. Most of this data was in English, and both the amount of training data and performance for each language varied considerably.

Since fine-tuning models may lead to catastrophic forgetting, research has looked toward non-parametric methods to improve performance. One such method is
token-level $k$NN search,
first introduced by \citet{khandelwal2020kNNLG} for language generation, and then applied to machine translation (MT) by \citet{khandelwal2021kNNMT}.
This $k$NN method, described in detail in \cref{sec:knn},
involves storing the hidden states together with each token in a sequence as key-value pairs in
an optimized structure called a \emph{datastore}.
At inference time, at each step, the model's hidden state is used to search the datastore for the $k$ nearest tokens, and the
output probability of the found tokens is
changed.
One
benefit of $k$NN is that one can create separate datastores depending on any category that one wishes to adapt a model's performance to, without needing to fine-tune the model. Although the datastores require space on the disk, they tend to be smaller than the weights that would otherwise need to be stored for a fine-tuned model, especially for very big models.
Additionally, we find $k$NN promising as it is a rare successful departure from the dominant paradigm of parametric linear classification heads.

Compared to text-to-text language generation models, speech introduces a new variable, namely pronunciation variability, which influences ASR performance both for individual speakers, as well as for speaker groups, which has the potential to introduce new forms of bias.

Our contributions are extending $k$NN at a token level to the ASR task, assessing the viability of $k$NN for speaker adaptation, as well as assessing Whisper's bias in Dutch for gender, accent, and age, and how $k$NN impacts it.

\section{Background and Related Work}
\subsection{Automatic Speech Recognition}
Whisper \citep{radford2023whisper} is a multilingual transformer end-to-end ASR model. Initially, all Whisper sizes were trained on over 680,000 hours of supervised speech data. Subsequent versions of the large model (large-v2 and large-v3) are trained on more data, including weakly and pseudo-labeled audio. This sets it apart from other models that are partially trained in an unsupervised manner, such as Wav2Vec 2.0 \citep{baevski2020wav2vec2}, opening up the possibility of transferring techniques previously applied to transformer encoder-decoder text-to-text language generation models.

\subsection{$k$NN}\label{sec:knn}
Both speech and language models tend to fall short outside of the domains that they were trained on
Fine-tuning them on these tasks, however, is expensive, and may lead to (catastrophic) forgetting \citep{dingliwal2022LLMdomainprompt}. \citet{khandelwal2020kNNLG, khandelwal2021kNNMT} propose a non-parametric method for adjusting the model output, namely $k$NN at a token-level for natural language generation (NLG) and machine translation (MT). They create a datastore with hidden states generated from the input and decoded output so far as keys and reference output tokens as values. At inference time, the hidden state for an input and output generated so far is used as a query $q$ to search the datastore for the $k$ nearest neighbors. The probability for each neighbor is obtained by
\begin{equation}
    p(k_i) \propto \text{exp}(-d_i),
\end{equation}
where $d$ represents the distance between the query $q$ and each neighbor. The probabilities of all non-unique tokens are then summed over:
\begin{equation}
    p_{k\text{NN}}(y) = \sum_i\mathds{1}_{y=v_i}p(k_i).
\end{equation}
Finally, this vocabulary distribution is interpolated with the original model's distribution:
\begin{equation}
    p(y) = \lambda p_{k\text{NN}}(y) + (1 - \lambda) p_{\text{model}}(y).
\end{equation}

In NLG and MT,
\citet{khandelwal2020kNNLG, khandelwal2021kNNMT} show that
$k$NN improves over
no $k$NN, and sometimes even over a fine-tuned model in both in-domain and out-of-domain settings, while needing fewer resources to train the datastore. A downside, however, is the increase in decoding time, as a $k$NN search needs to be done at each decoding step.
This computational expense depends on the size of the datastore and
several mitigating strategies have been studied, including
the use of smaller datastores \citep{dai2023simple},
optimized data structures for approximate neighbor
lookups \citep{johnson2021faiss}, or chunked lookups \citep{martins-etal-2022-chunk}.

\subsection{Augmenting ASR}
Both nonretrieval- as well as retrieval-based approaches have been studied in the context of ASR.

\citet{dingliwal2022LLMdomainprompt} adapt a large language model (LLM) that reranks ASR hypotheses. They do so by adding embedding parameters (``domain prompts'') to the embedding layer of the LLM that they train for a specific domain.

\citet{chan2023acousticcatalog} use a combination of $k$NN and attention fusion. Their datastore contains audio embeddings as keys, and text embeddings as values. These are then used directly in a cross-attention fusion layer to get the final vocabulary distribution per token.
\citet{mittal2023kNN} create a hidden-state trie that they search over with $k$NN. Their method ensures that only words that exist in the trie can be output. \citet{wang2024whisperICL} use in-context learning by providing the $k$ nearest audio and transcript as prompts to Whisper. The approach by \citet{sarı2020speakerAdapt} attends over speaker i-vectors (identity vectors) in the memory for speaker adaptation.

\subsection{Bias in ASR}
Bias is found across various categories in ASR. It has been found that in some cases, speech models perform better for women \citep{koenecke2020racialDisparities}, while in others, they perform worse \citep{garnerin2019frenchBias}. Performance also varies by age and accent \citep{feng2021dutchBias, feng2024dutchMandarinBias, fuckner2023dutchBiasW2vWhisper}, as well as race \citep{koenecke2020racialDisparities}, where children, older adults, speakers with a ``non-standard'' accent, and Black speakers tend to be impacted negatively. In consequence, these groups have less access to accessibility tools, such as voice assistants, and other services that use ASR in the pipeline.

\section{Methods}

\paragraph{Datasets}
We use four datasets for our experiments: VoxPopuli \citep{wang-etal-2021-voxpopuli}, LibriSpeech \citep{panayotov2015librispeech}, CommonVoice
\citep{ardila-etal-2020-common}
and RixVox \citep{rekathati2023rixvox}. LibriSpeech is an English-spoken dataset and RixVox is a Swedish dataset. For VoxPopuli we use the English portion of the dataset, and for CommonVoice we use the Dutch portion of version 18.0.

\paragraph{Models}
We perform all experiments on OpenAI's Whisper speech model \citep{radford2023whisper}. To assess the effect of $k$NN on different model sizes and to find the optimal settings, we run a hyperparameter search on the VoxPopuli dataset using the Whisper tiny, medium, and large-v3 models. For the remainder of the datasets, we only tune $\lambda$ on Whisper large-v3. We used the FAISS library \citep{johnson2021faiss} to build the datastores. We used the IVFPQ index with 2048 centroids, code size 64, and 32 probes or partitions.

\paragraph{Tuning the $k$NN Hyperparameters}
For our choice of hyperparameters, we follow the setup by \citet{martins2023empiricalkNN} for $\lambda$, $T$, and $k$. Thus: $\lambda \in \{0.3, 0.4, 0.5, 0.6\}$, \( T \in \{1,10,100\}\), and $k \in \{4,8,16\}$. For the three different model sizes and the VoxPopuli dataset, we tune on the full range of these parameters. For the remaining datasets, we choose the $T$ and $k$ found to be optimal for Whisper large-v3 and only tune $\lambda$. The results are based on the best hyperparameter configuration for each dataset and model combination described above. Any ties are broken at random.

\paragraph{Speaker Adaptation}\label{sec:method_speaker_adaptation}
To assess the usefulness of $k$NN for speaker adaptation, we build personalized datastores for a random subset of 33 speakers in the RixVox datasets. Each datastore only included the hidden states and tokens obtained for each respective speaker. We compare this to building a datastore of the same size for each speaker, but instead filled with random tokens uniformly sampled from the full datastore. For both speaker-level conditions, $\lambda$ was tuned for each speaker.

\paragraph{Bias}
We use the CommonVoice dataset to test for bias related to gender, accent, and age before and after applying $k$NN. Each of these characteristics are self-reported by the speakers.\footnote{Gender is not binary, but our analysis is limited by the dataset only including the qualifiers \texttt{female\_feminine}, \texttt{male\_masculine}, or no qualifier. Additionally, some recordings were associated with multiple accents. We always picked the first one.} Since not all rows were marked with this information, for each category only a subset of the data was examined.

\section{Results}
\subsection{Main Results}
\begin{table}[] \small
    \centering
    \begin{tabular}{llrr}
    \toprule
        \multirow{2}{*}{Whisper} & \multirow{2}{*}{Dataset} & \multicolumn{2}{c}{WER}  \\
          &  & Vanilla & $k$NN \\
    \midrule
        tiny & \multirow{3}{*}{voxpopuli.en} & 12.28 & 12.28 \\
        medium &  & 7.48 & 6.90 \\
        large-v3 &  & 8.13 & 7.31 \\
    \midrule
        \multirow{4}{*}{large-v3} & LS-clean & 1.89 & 1.95 \\
          & LS-other & 3.79 & 3.93 \\
          & CV & 4.58 & 4.38 \\
          & RixVox & 16.7 & 14.54 \\
    \bottomrule
    \end{tabular}
    \caption{Overall WER without and with $k$NN.}
    \label{tab:main_kNN}
\end{table}

Table \ref{tab:main_kNN} shows the main results,
while hyperparameter tuning results are deferred to \cref{sec:hyperparam}.
We report the word error rate (WER, \citealp{vogel2000WER}), \ie, the number of mistakes divided by the total number of words in the reference, calculated using JiWER\footnote{\url{https://jitsi.github.io/jiwer}}.
On VoxPopuli, $k$NN
improves the results for the medium and large-v3 models, while having no effect on the tiny model. For LibriSpeech, we observe a decrease in performance, while observing an increase for both CommonVoice and RixVox. For LibriSpeech, we observe a decrease in performance, while observing an increase for both CommonVoice and RixVox. For LibriSpeech, with such a strong performance, there is little that can be improved on with a $k$NN approach. Manual inspection of the errors reveals they are mostly difficult or ambiguous cases, including names, written out accents, old words, or other minor differences in spelling, interjections, or compounds. The other datasets may be noisier or Whisper may be less familiar with them, which could mean there is more room for improvement.

Table \ref{tab:example_neighbours} shows an example of the nearest neighbors found for a sentence in CommonVoice. Some of the prior contexts of the neighbors include the current decoded context. We also see that some of the following words are the same as the word that was decoded next, and otherwise includes the beginning of the next word. This suggests that the hidden states do not only encode information about the current token, but also about the surrounding tokens.

\begin{table}[] \small
    \centering
    \begin{tabular}{ll}
    \toprule
        (...) de voornaamste punten \textcolor{examplered}{willen} aanstippen. \\
    \midrule
        (...) belangrijke punten \textcolor{examplered}{willen} aanstippen. \\
        (...) kwestie die ik zou \textcolor{examplered}{willen} aanstippen, (...) \\
        Ik zou graag drie punten \textcolor{examplered}{willen} aanhalen. \\
        Zou u mijn excuses \textcolor{examplered}{willen} aanvaarden. \\
    \bottomrule
    \end{tabular}
    \caption{Example neighbors for CommonVoice. The top row is the decoded sentence, the bottom four are the four neighbors found inside their original contexts.
    \textcolor{examplered}{Red} marks the current token being decoded.}
    \label{tab:example_neighbours}
\end{table}

\subsubsection{The Effect of $k$, $T$, and $\lambda$}
\begin{figure}
    \centering
    \includegraphics[width=\linewidth]{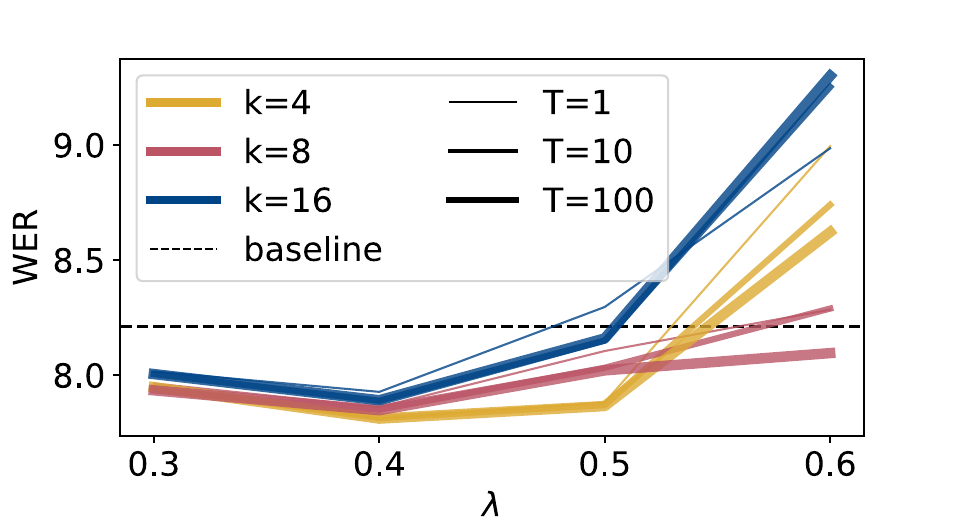}
    \caption{WERs on VoxPopuli.en dev with Whisper large-v3 for different $k$s (line color), $T$s (line width), and $\lambda$ (x-axis).}
    \label{fig:dev-vp}
\end{figure}

Figure \ref{fig:dev-vp} shows how $k$, $T$, and $\lambda$ affect the WER on the VoxPopuli dev set with Whisper large-v3.

We observe the following patterns: First, results generally improve as we reduce the number of neighbors. Only for $\lambda=0.6$ do we get the best results for $k=8$. Second, setting $T$ higher tends to give better results, but this is also up to variability depending on the other hyperparameters. Finally, we find that, generally, we get the best results at $\lambda=0.4$.

These results contrast with those obtained by \citet{khandelwal2020kNNLG, khandelwal2021kNNMT}. They find that a higher $k$ tends to improve performance, whereas we find the opposite trend. It could be that for ASR with a transformer encoder-decoder, adding more neighbors only increases the chance of retrieving irrelevant neighbors. Second, we find that the optimal $\lambda$ is also different from theirs. They find a lower $\lambda$ for in-domain $k$NN, and a higher one for out-of-domain $k$NN. In this work, Whisper has been trained on a large variety of data, very likely including (some of) the datasets used in this work. Thus, applying $k$NN with VoxPopuli is somewhere between in-domain and out-of-domain tuning. This suggests that the more out-of-domain the dataset used for $k$NN, the higher $\lambda$ should be, as the speech (or language) model might be more unfamiliar with the patterns in the target text. This is further supported by needing $\lambda=0.5$ for RixVox, whereas $\lambda=0.4$ was best in all other cases. As RixVox was released in 2023, it is less likely, but not impossible, for Whisper large-v3 to be trained on it.

\subsection{Speaker Adaptation}
In this section we analyze the results for the speaker adaptation experiments. The results can be found in Table \ref{tab:speaker_adaptation}. Using the complete datastore results in a larger average improvement per speaker compared to any of the other methods. Using a personal datastore leads to smaller improvements and is comparable to using the random datastore of equal size (Section \cref{sec:method_speaker_adaptation}). What the smaller datastores lose in accuracy they make up for in efficiency: with the full datastore, a single file takes on average $\sim$1 minute and 15 seconds to transcribe, while for the smaller datastores it only takes roughly 7 seconds. From a private-user perspective, this win in time could be the deciding factor, if it means being able to use some transcription technology or voice assistant live.

\begin{table}[] \small
    \centering
    \begin{tabular}{lrrrr}
    \toprule
        Setting & Van. & Rand. & Pers. & Gen. \\
    \midrule
        Mean & 16.09 & 15.94 & 15.69 & 13.87 \\
        Std. & 2.69 & 2.53 & 2.64 & 2.78 \\
    \bottomrule
    \end{tabular}
    \caption{WERs for speaker adaptation on RixVox with different settings. ``Van.'' refers to no $k$NN, ``Rand.'' and ``Pers.'' are random and personal datastores respectively with a tuned $\lambda$ per speaker, and ``Gen.'' indicates the full datastore.}
    \label{tab:speaker_adaptation}
\end{table}

\subsection{Bias}
\begin{figure}
    \centering
    \includegraphics[width=\linewidth]{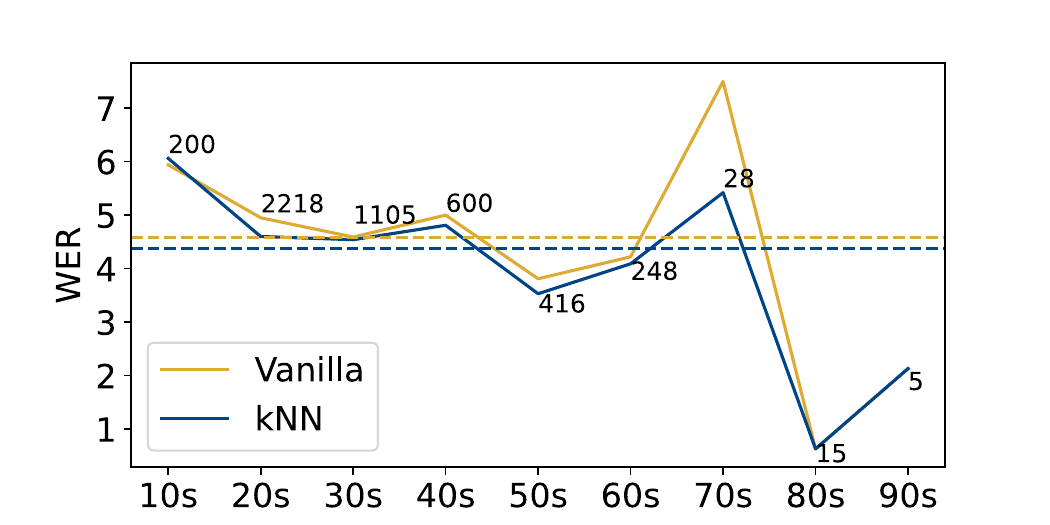}
    \caption{WER per age group for CommonVoice 18.0 NL using Whisper large-v3. The horizontal dashed lines represent the overall results without and with $k$NN. The numbers in the graph indicate the
bin count.%
}
    \label{fig:age}
\end{figure}

In this section we analyze the transcription performance with and without $k$NN on various speaker groups in the CommonVoice dataset. We cover three categories: gender, accent, and age. See Table \ref{tab:CV_bias} and Figure \ref{fig:age} for the results.

First, for gender, we see that Whisper performs comparably for both, but with $k$NN leads to a larger improvement for women than for men.

Second, Dutch speakers from the Netherlands are recognized better than those from Belgium. The WER improves for speakers from both groups, with Belgian speakers benefiting somewhat more. It is difficult to ascertain whether this is due to accent or vocabulary, as the dataset contains Dutch vocabulary from both Belgium and the Netherlands.

When it comes to age, Whisper performs the worst on people in their teens and seventies, and best on those in their fifties\footnote{The eighties and nineties group are each only represented by one speaker, and have been included for completeness.}. No other clear patterns can be observed. Adding $k$NN results in a nearly consistent improvement, except for teenagers, for whom the WER increases.

The accent result cannot be explained by the number of recordings in the training data for $k$NN, as there are approximately twenty times more recordings of Dutch than Belgian speakers. For gender and age, this is inconclusive, as approximately half the train recordings contain no information on this.

\begin{table}[] \small
    \centering
    \begin{tabular}{lrrr}
    \toprule
        Category & Vanilla & $k$NN & \#Test recordings \\
    \midrule
        Overall & 4.58 & 4.38 & 11309 \\
    \midrule
        Women & 4.69 & 4.27 & 1048 \\
        Men & 4.70 & 4.55 & 3618 \\
    \midrule
        Netherlands & 4.47 & 4.30 & 3505 \\
        Belgium & 5.30 & 4.97 & 1038 \\
    \bottomrule
    \end{tabular}
    \caption{WER across different categories in CommonVoice 18.0 (NL) with Whisper large-v3.}
    \label{tab:CV_bias}
\end{table}

\section{Conclusion}
In this paper, we find that $k$NN can improve the ASR performance for Whisper, a transformer end-to-end speech model. Additionally, we observe that using a smaller datastore for individual speakers can still lead to an improvement, trading in the performance from using a full datastore for speed. Finally, we find that Whisper's performance in Dutch is similar for women and men, is worse for Belgian speakers of Dutch than for Dutch speakers, and varies by age. Applying $k$NN shows larger improvements for women and Belgian speakers, and leads to improvements for most age groups, except for teens. Our analysis suggests some improvements from $k$NN seem to stem from Whisper's decoder's predictive nature, as the context of some of the retrieved neighbors also includes the same decoded continuation.

\section{Limitations}
In this study, we used a subset of speakers for the speaker adaptation experiment. It is possible that, when using the full dataset, we could have seen a different pattern. Furthermore, due to time-constraints, we fine-tune $\lambda$, and fix $k$ and $T$ for the other datasets besides VoxPopuli, which may have led to not observing the full picture for the effect of $k$NN in different languages and settings. It is also difficult to tell to what extent Whisper is or is not familiar with the data used in this study, as its training data has not been made public. Additionally, all of the languages included in this study are Indo-European of origin, resulting in certain similarities, such as overlapping vocabulary due to cognates, as well as all having a synthetic morphology. They also (mostly) use the same alphabet. More work is needed to see the impact of $k$NN on languages with other typological features and/or from different language families. For the other datasets besides CommonVoice, more work is needed to assess whether $k$NN affects all speaker groups equally. Finally, in this work, we took bias to mean unequal performance across the given categories and labels in the dataset. However, for each of the categories a more complete analysis is needed, as gender was only analyzed as binary, there are more accents of Dutch than the ones described, and there was no data for children's speech.

\section*{Acknowledgments}
The authors thank the members of the TAIM lab and the UvA LTL for helpful suggestions.
This publication is part of the project \emph{ROBUST: Trustworthy AI-based Systems for Sustainable Growth}
with project number KICH3.LTP.20.006, which is (partly) financed by the Dutch Research Council
(NWO), RTL, and the Dutch Ministry of Economic Affairs and Climate Policy (EZK) under the program
LTP KIC 2020-2023. VN is partly funded by the Dutch Research Council (NWO) via VI.Veni.212.228
and by European Union's Horizon Europe research and innovation programme
via UTTER 101070631.

\bibliography{kNN}

\appendix

\section{RixVox Splits}
We created a new RixVox split for the speaker adaptation experiments. This split was also used for the general RixVox experiments. Table \ref{tab:rixvox_data} shows the new data distribution.
\begin{table}[h] \small
    \centering
    \begin{tabular}{l r r r}
    \toprule
    Split & train & dev & test \\
    \midrule
    Hours & 1436 & 30 & 30 \\
    N.~speeches & 212673 & 4428 & 4500 \\
    Min speeches per speaker & 964 & 20 & 21 \\
    Max speeches per speaker & 5419 & 113 & 113 \\
    \bottomrule
    \end{tabular}
    \caption{RixVox split information.}
    \label{tab:rixvox_data}
\end{table}

\section{Optimal Hyperparameters}\label{sec:hyperparam}
Table \ref{tab:optimal_hyperparameters} reports the hyperparameters for our main experiments. 
\begin{table}[] \small
    \centering
    \begin{tabular}{llrrr}
    \toprule
        Model name & Dataset & $k$ & $T$ & $\lambda$ \\
    \midrule
        tiny & \multirow{3}{*}{VoxPopuli.en} & 4 & 10 & 0.4 \\
        medium &  & 16 & 100 & 0.4 \\
        large &  & 4 & 100 & 0.4 \\
    \midrule
        \multirow{3}{*}{large-v3} & LibriSpeech & 4 & 100 & 0.4 \\
         & CommonVoice 18.0 NL & 4 & 100 & 0.4 \\
         & RixVox & 4 & 100 & 0.5 \\
    \bottomrule
    \end{tabular}
    \caption{Optimal hyperparameters for different Whisper sizes and datasets.}
    \label{tab:optimal_hyperparameters}
\end{table}

\section{Use of AI Assistant}
We used GitHub Copilot\footnote{https://github.com/features/copilot} for auto-completing single lines of code that were repetitive with respect to the rest of our code.

\section{Compute}
For this project, we used approximately 5000 GPU hours. This includes restarts, experiments that were not included, as well as runs leading to the final output. These hours are spread out over the following GPUs: NVIDIA GeForce GTX TITAN X, NVIDIA L40, NVIDIA RTX A6000, NVIDIA TITAN X (Pascal), and Tesla P40.

\section{Code}
The code for this project can be found here: \url{https://github.com/MKNachesa/kNN}.
\end{document}